\documentclass[letterpaper, 10 pt, conference]{ieeeconf}  

\IEEEoverridecommandlockouts                              

\usepackage{graphics} 
\usepackage{graphicx}
\usepackage{multirow}
\usepackage[utf8]{inputenc}
\usepackage{mathptmx} 
\usepackage{times} 
\usepackage{amsmath} 
\usepackage{amssymb}  
\usepackage{siunitx}
\usepackage{blindtext}
\usepackage{booktabs}
\usepackage{lipsum}
\usepackage{pifont} 

\usepackage{xcolor}

\usepackage{pgfplots}
\usepgfplotslibrary{groupplots}

\usepackage{tikz}
\usetikzlibrary{calc}
\usetikzlibrary{arrows, arrows.meta}
\usetikzlibrary{shapes,fit}
\usetikzlibrary{positioning}
\usetikzlibrary{shadings}
\usetikzlibrary{shadows.blur}
\usepackage[T1]{fontenc}

\usepackage{braket}

\DeclareMathAlphabet\mathcal{OMS}{cmsy}{m}{n}
\SetMathAlphabet\mathcal{bold}{OMS}{cmsy}{b}{n}

\title{\LARGE \bf Improved Multi-Scale Grid Rendering of Point Clouds for Radar Object Detection Networks}

\author{Daniel K\"ohler$^{1,2}$, Maurice Quach$^{3}$, Michael Ulrich$^{3}$, Frank Meinl$^{1}$, Bastian Bischoff$^{3}$ and Holger Blume$^{2}$
\thanks{$^{1}$Daniel K\"ohler and Frank Meinl are with Robert Bosch GmbH, Cross-Domain Computing Solutions, Germany\newline
        {\tt\small daniel.koehler2@de.bosch.com}}
\thanks{$^{2}$Daniel K\"ohler and Holger Blume are with Leibniz University Hannover, Institute of Microelectronic Systems, Germany}%
\thanks{$^{3}$Maurice Quach, Michael Ulrich and Bastian Bischoff are with Robert Bosch GmbH, Corporate Research, Germany}%
}

\usepackage{xspace}

\makeatletter
\DeclareRobustCommand\onedot{\futurelet\@let@token\@onedot}
\def\@onedot{\ifx\@let@token.\else.\null\fi\xspace}

\def\eg{e.g\onedot} 
\def\ie{i.e\onedot}

\def\wrt{w.r.t\onedot} 

\def\no{no\onedot}
\makeatother

\usepackage{cite}

\newcommand{\fig}[1]{Fig.\,\ref{#1}}

\newcommand{\tab}[1]{Tab.\,\ref{#1}}

\usepackage{caption}
\usepackage{ragged2e}
\newcommand{\figuredesc}[1]{
  \begingroup
  \par
  \justifying\small
  \noindent #1
  \par
  \endgroup}

\begin{document}

\maketitle
\thispagestyle{empty}
\pagestyle{empty}

\begin{abstract}

Architectures that first convert point clouds to a grid representation and then apply convolutional neural networks achieve good performance for radar-based object detection.
However, the transfer from irregular point cloud data to a dense grid structure is often associated with a loss of information, due to the discretization and aggregation of points. 
In this paper, we propose a novel architecture, multi-scale KPPillarsBEV, that aims to mitigate the negative effects of grid rendering.
Specifically, we propose a novel grid rendering method, KPBEV, which leverages the descriptive power of kernel point convolutions to improve the encoding of local point cloud contexts during grid rendering.
In addition, we propose a general multi-scale grid rendering formulation to incorporate multi-scale feature maps into convolutional backbones of detection networks with arbitrary grid rendering methods.
We perform extensive experiments on the nuScenes dataset and evaluate the methods in terms of detection performance and computational complexity.
The proposed multi-scale KPPillarsBEV architecture outperforms the baseline by 5.37\% and the previous state of the art by 2.88\% in Car AP4.0 (average precision for a matching threshold of 4 meters) on the nuScenes validation set.
Moreover, the proposed single-scale KPBEV grid rendering improves the Car AP4.0 by 2.90\% over the baseline while maintaining the same inference speed.

\end{abstract}

\section{Introduction}

Accurate environmental perception, \eg detection of road users, is crucial for safe and reliable operation of autonomous driving platforms.
In this context, radar sensors play an increasingly important role in addition to camera and lidar, as they are cost effective and highly robust against adverse weather conditions.
Furthermore, radar perception systems are improving at a rapid pace and promise great potential \wrt upcoming high-resolution radar sensors that provide much denser point clouds.

\begin{figure}[tb]
	\centering
	\input{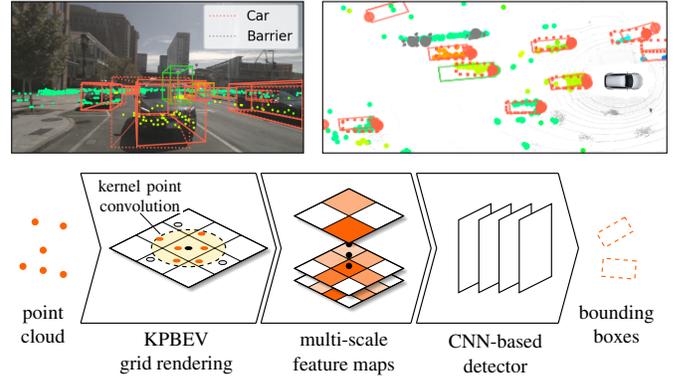}
	\caption{Object detectors based on CNNs render point clouds into a discrete grid representation, which can lead to information loss. We propose the novel multi-scale KPPillarsBEV architecture and improve grid rendering with two key methods. First, we propose KPBEV, a new grid rendering module, that uses kernel point convolutions to learn more expressive features from the radar point cloud. Furthermore, we incorporate scale-aware features into grid-based detectors with a general multi-scale grid rendering formulation.}
	\label{fig:teaser}
\end{figure}

With regard to object detection, previous work in lidar \cite{Chen2017,Zhou2019,Yang2019,Lang2019} and radar \cite{Xu2021,Scheiner2021,Ulrich2022} domains has shown that grid-based architectures outperform methods that operate directly on the point cloud.
Typically, such grid-based architectures first project point clouds into a 2D bird's-eye view (BEV) or 3D voxel representation and then apply convolutional neural networks (CNNs).
In such architectures, the transition from continuous point cloud to discrete grid representation is an essential step.
The grid resolution has to be parametrized carefully, as it has a strong influence on both detection performance and computational complexity.
A high grid resolution allows to learn fine-grained features from the input points. 
On the other hand, it can inhibit the propagation of information over larger spatial extents and increases computational complexity.
Specifically, the complexity scales quadratically when increasing the resolution of a 2D grid.
In contrast, a low grid resolution improves inference speed and reduces memory requirements of the model.
However, it can lead to information loss during grid rendering, as more points fall into the same grid cell, which makes cell-wise feature aggregation more difficult.

In grid-based architectures, information exchange across multiple cells is typically performed by convolutional layers after discretizing the point cloud to a grid.
However, finer geometric details in the point cloud may already be lost at this stage.
In \cite{Ulrich2022}, this problem is addressed with hybrid models that combine grid-based methods with point-based preprocessing consisting of a graph neural network (GNN) \cite{Shi2020} or kernel point convolutions (KPConvs) \cite{Thomas2019}.
Still, information is lost during grid rendering which is detrimental to object detection performance.
To address this issue, we propose KPBEV, a novel grid rendering module that aims to improve information flow between points within each cell and across multiple cells by directly integrating point convolutions into grid rendering.

Furthermore, traffic participants and other objects of interest can have very different extents, \eg trucks in comparison to pedestrians. 
This challenge is commonly addressed by using detection backbones that down- \cite{He2016} and upsample spatial features \cite{Lin2017}.
In the context of lidar object detection there are different multi-scale grid rendering approaches \cite{Ye2020,Song2020a,Kuang2020}, that directly process the point cloud into feature maps of different spatial scales.
In this paper, we propose a general formulation for multi-scale grid rendering which is agnostic to the actual grid rendering method.
We also extensively evaluate the benefit of multi-scale grid rendering compared to single-scale with different architectures and grid rendering methods.


Finally, we propose a novel architecture, multi-scale KPPillarsBEV, that brings together our novel KPBEV grid rendering and multi-scale grid rendering formulation with point-based preprocessing \cite{Ulrich2022}.
KPPillarsBEV outperforms the current state-of-the-art radar object detection models.
In addition, variations of this architecture offer outstanding tradeoffs between detection performance and computational complexity.

To summarize, the contributions of this work are the following
\begin{itemize}
	\item We propose KPBEV, a grid-rendering module that leverages KPConvs to improve the information exchange between input points and mitigate discretization effects.
	\item We propose a novel multi-scale grid rendering formulation to incorporate multi-scale feature maps rendered directly from the point cloud into detection networks.
	\item We propose a novel multi-scale KPPillarsBEV architecture, that brings together KPBEV, multi-scale rendering and point-based preprocessing, which outperforms the state-of-the-art for radar object detection.
	\item We evaluate detection performance and computational complexity of this architecture and its variants on nuScenes \cite{Caesar2020}. We find that these variations provide outstanding tradeoffs between performance and complexity compared to the state of the art.
\end{itemize}
\section{Related Work}

\subsection{Radar Object Detection}

Data-driven methods for radar-based object detection can be divided into two categories: spectrum-based and point cloud based.
Spectrum-based object detection networks use dense representations of the spectral radar data that can include range, radial velocity (Doppler), azimuth and elevation dimensions.
For this purpose, CNNs are applied to different 2D projections of the radar tensor, such as range-Doppler and \cite{Brodeski2019,Rebut2022} range-azimuth maps \cite{Lim2019,Wang2021a} or combinations thereof \cite{Major2019}.
In GTRNet \cite{Meyer2021} and T-RODNet \cite{Jiang2023}, convolutional feature extractors are combined with graph- and attention-based methods, respectively.
On the one hand, using raw spectral data avoids loss of information due to downstream steps in the radar signal processing chain, \eg constant false alarm rate detectors.
On the other hand, radar spectra approaches are computationally more complex and are more memory-intensive and sensitive to the sensor configuration, which can impede generalization of neural networks based methods.

Thus, a second research direction focuses on the more abstract and sensor-independent point cloud representation of radar data.
This can be further subdivided into point- and grid-based methods.
Building on the pioneering architectures of PointNet \cite{Qi2017a} and PointNet++ \cite{Qi2017}, point-based models focus on directly aggregating information from points clouds, which are often irregular and sparse.
In the radar domain, such approaches were applied to classification \cite{Ulrich2021}, semantic segmentation \cite{Schumann2018} and object detection \cite{Danzer2019,Bansal2020,Scheiner2021}.
Recent work also investigated more complex point-wise feature extractors such as graph and kernel point convolutions \cite{Svenningsson2021,Nobis2021,Ulrich2022} or transformers \cite{Bai2021,Zeller2023}.
In contrast, grid-based methods first project the point cloud into a regular grid in order to leverage the capability of CNNs to extract spatial features.
To this end, \cite{Dreher2020,Scheiner2021} use variations of the YOLO architecture \cite{Redmon2016} or feature pyramid networks \cite{Meyer2019a,Lin2017} to detect objects in BEV projections of the radar point cloud.
In \cite{Scheiner2021,Xu2021,Tan2022,Palffy2022}, the PointPillars \cite{Lang2019} feature encoder is used to learn a more abstract grid representation of the point cloud.
\cite{Ulrich2022} proposes a hybrid architecture, combining grid-based methods with point-based preprocessing to learn more expressive features from point clouds and to improve the detection performance.

\subsection{Grid Rendering of Point Clouds}

Converting irregular and sparse point clouds into regular and dense grid representations is a common task for 3D object detection models that leverage CNNs for feature extraction. 
Methods for such grid rendering of point clouds can be distinguished by the resulting grid representation and the type of encoding used to obtain cell-wise features.

MV3D \cite{Chen2017} and PIXOR \cite{Yang2019} utilize handcrafted features such as intensity, density and height maps to obtain a BEV representation from point clouds that can be processed with 2D convolutions. 
In contrast, VoxelNet \cite{Zhou2017} and PointPillars \cite{Lang2019} rely on learnable feature extractors that apply simplified PointNets \cite{Qi2017a} to all points within the same grid cell.
Building on this approach, SECOND \cite{Yan2018} and PillarNet \cite{Shi2022} employ hierarchical encoders that successively downsample the initial feature map (along one or multiple spatial dimensions) with sparse convolutions before feeding it into a convolutional detection backbone. 

Besides the previously mentioned methods which render the point cloud into a feature map at a single scale, several methods generate feature maps at different spatial scales in parallel. 
For instance, attentive voxel encoding layers \cite{Ye2020} or a PointPillars variation \cite{Song2020a} are applied for different grid resolutions and the resulting grids are fused at a common scale.
Voxel-FPN \cite{Kuang2020} extends this by additionally forwarding the individual feature maps directly to the detection backbone.

\section{Method}

\subsection{Overview}

In this paper, we aim to improve information flow from irregular point cloud data to image-like, dense feature maps in grid-based object detection networks.
To this end, we propose KPBEV, a module that encodes point features into a grid using kernel point convolutions (KPConvs) \cite{Thomas2019}.
In addition, we propose a general multi-scale rendering formulation to encode the point cloud into grids with different spatial scales.
Finally, we propose a novel architecture, multi-scale KPPillarsBEV, that brings together KPBEV, multi-scale rendering and point-based preprocessing.

\subsection{KPBEV}

\begin{figure}[tb]
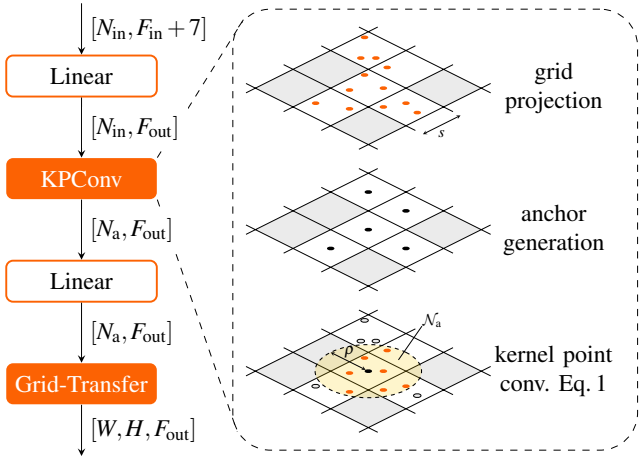

	\centering
	\resizebox{\columnwidth}{!}{%
		\begin{tikzpicture}[node distance=1.5cm]

\pgfdeclarelayer{background}
\pgfsetlayers{background,main}

\input{graphics/03_kpbev/left_side.tex}

\node (rightside) at ($(test.east|-test)+(3.7,0)$) {
	\input{graphics/03_kpbev/right_side.tex}
};

\begin{pgfonlayer}{background}
\node[draw, dashed, rounded corners=.55cm, fit=(rightside),inner sep=0mm] (box) {};
\draw[dashed] (l1.north east) -- ($(box.north west)+(0.293*0.55cm, -0.293*0.55cm)$);
\draw[dashed] (l1.south east) -- ($(box.south west)+(0.13*0.55cm, 0.5*0.55cm)$);

\end{pgfonlayer}

\end{tikzpicture}
	}%
	\caption{KPBEV extracts features at each anchor point, \ie the centers of non-empty grid cells, using rigid KPConvs. Features are transformed by a linear layer before and after this step. Finally, the features are transferred to the grid leveraging the one-to-one correspondence between anchor points and grid cells. Note that batch normalization and ReLU activation are applied after each layer.}
	\label{fig:kpbev}
\end{figure}

In previous work \cite{Lang2019, Zhou2017}, cell-wise features are obtained by aggregating over points within the same cell using a pooling function.
Local neighborhood features across multiple cells are then encoded by applying convolutions.
However, at this stage, fine-grained details of the input point cloud may already be lost due to discretization and aggregation.

Thus, we propose a novel grid-rendering module, KPBEV, that aims to improve information flow between points in order to learn more expressive features of local neighborhoods.
For this purpose, it exploits both the descriptive power of KPConvs and the flexible parametrization of the receptive field for cell-wise feature aggregation.
For each non-empty grid cell (anchor point), we aggregate information from points in its neighborhood using a KPConv.

In previous approaches, a point is usually assigned to a single cell, whereas KPBEV can use the information from a point in multiple cells.
In other words, KPBEV can include points from adjacent cells which enables it to better capture local context.
Moreover, the anchor-wise aggregated features can be directly transferred to the grid, as there is a one-to-one mapping between each anchor point and each non-empty grid cell.

The structure of the KPBEV module is depicted in \fig{fig:kpbev}.
The input is a point cloud $\mathcal{P}_\textnormal{in}=\set{x_0,\ldots,x_{N_\textnormal{in}}}$ of $N_\textnormal{in}$ points $x_i\in \mathbb{R}^2$ and their corresponding features $f_i\in\mathbb{R}^{F_\textnormal{in}}$ in $\mathcal{F}_\textnormal{in}=\set{f_0,\ldots,f_{N_\textnormal{in}}}$.
First, the input points are projected into a grid of width $W$, height $H$ and spatial scale $s$.
Based on this grid projection, a new set of $N_\textnormal{a}$ points is created, that contains a point at the center of each non-empty grid cell.
These points serve as anchors for encoding cell-wise contexts from neighboring input points $\mathcal{N}_\textnormal{a} = \set{x_i \in \mathcal{P}_\textnormal{in}\;|\; \left\lVert x_i-x_a\right\rVert \leq \rho}$ within a specified radius $\rho$.

To this end, the input features $f_i$ are enhanced with additional features $\set{(x_i-x_a), (x_i-x_c), x_c, n}$ where $(x-y)$ refers to the relative position of two points $x$ and $y$, $x_a$ is the corresponding anchor point of $x_i$ and $x_c$ is the centroid of the $n$ points within the same grid cell.
The resulting augmented features $f_i\in\mathbb{R}^{F_\textnormal{in}+7}$ are then processed with a linear layer to obtain a more abstract representation $f'_i\in\mathbb{R}^{F_\textnormal{out}}$.

Subsequently, features are aggregated at each anchor point by applying a rigid KPConv.
KPConvs encode features at a given output coordinate from features of input points in a spherical neighborhood.
This is done via a set of kernel points $x_k$ that carry learnable weight matrices $W_k$ and are evenly distributed around the output point.
Specifically, the features of input points are processed by multiplication with the weight matrix of the respective kernel point and a distance-based correlation function $h(\cdot,\cdot)$.

In general, KPConvs and point convolutions allow for the set of input and output points to be different.
KPBEV takes advantage of this fact and aggregates features $f_{a}\in \mathbb{R}^{F_\textnormal{out}}$ at each anchor point $x_a$ from the respective neighboring input points $x_i$ and their abstract features $f'_i$
\begin{equation}
\label{eq:kpconv}
f_{a} = \sum_{x_i \in \mathcal{N}_a} f'_i \sum_k h(x_k,x_i-x_a) W_k
\end{equation}
with the linear correlation function
\begin{equation}
h(x_k, y) = \max\left(0,\; 1-\frac{\lVert x_k-y \rVert}{\rho_k}\right)
\end{equation}
where $\rho_k=\rho/2.5$ is the influence radius of each kernel point (see \cite{Thomas2019}). 
Note that, depending on the convolution radius $\rho$, the neighborhood context considered at each anchor may span across several cells. 
Consequently, information from points in adjacent cells can also be included at this stage to capture the local point cloud context.
In addition, KPConvs only have to be applied to each anchor point rather than each input point, which can lead to a significant reduction of computations, as we show in our experiments (see Sec.\,\ref{sec:experiments}).

The resulting anchor-wise features are once again passed through a linear layer. 
Since there is a one-to-one correspondence between anchor points and grid cells, the information aggregated at each anchor point can be directly transferred to the underlying grid to obtain a dense feature map $\mathcal{G}\in\mathbb{R}^{W\times H\times F_\textnormal{out}}$.
Although we illustrate and focus on the generation of 2D BEV representations (see \fig{fig:kpbev}), the proposed method can also be used to generate 3D voxel grids.
Furthermore, we focus on kernel point convolutions as they typically perform well in radar object detection \cite{Ulrich2022}.
However, it is also possible to combine the proposed method with other types of point-based convolutions, such as graph convolutions.

\subsection{Multi-Scale Grid Rendering}

\begin{figure}[tb]
	\centering
	\input{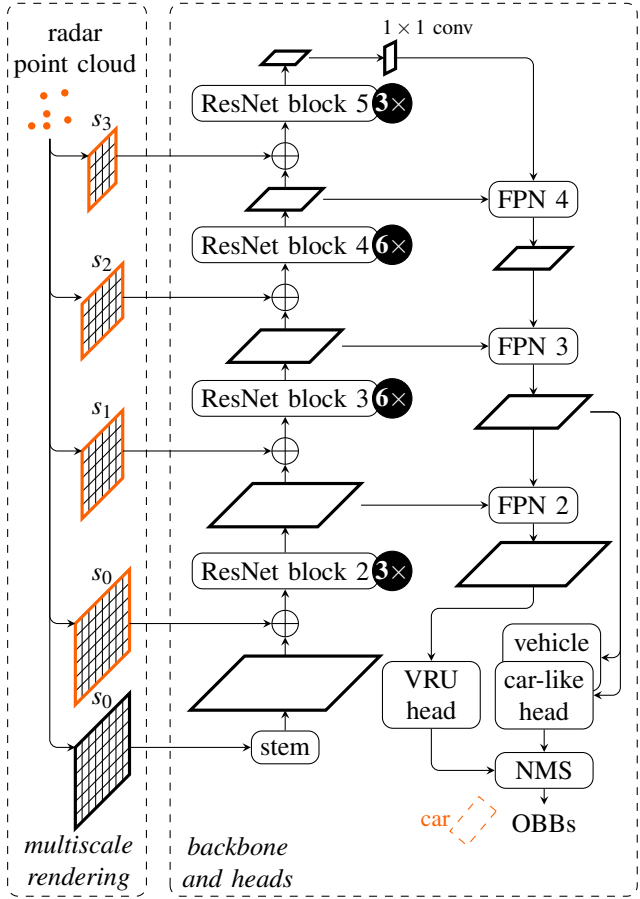}
	\caption{Structure of the multi-scale grid rendering in our network architecture. Given a grid rendering method, for each spatial scale $s_i$ in the backbone, the radar point cloud is rendered into a grid of corresponding scale (in orange). The resulting scale-aware features are fused at the respective backbone stage via concatenation $\oplus$. Finally, convolutional detection heads and non-maximum suppression (NMS) are applied to generate and filter oriented bounding boxes (OBBs) for different class groups. The architecture is a combination of a ResNet backbone network and a feature pyramid network (FPN).}
	\label{fig:network_architecture}
\end{figure}

Encoding point clouds into grid representations at different spatial scales in order to learn scale-aware features has been shown to be beneficial for lidar object detection \cite{Ye2020,Song2020a,Kuang2020}.
We propose a general multi-scale grid rendering formulation to learn scale-aware features.

Our approach differs from previous work in several points.
First, we propose a general approach that can be applied to any grid rendering method, \eg the PointPillars encoder \cite{Lang2019} and our newly proposed KPBEV encoder.
Second, our multi-scale approach is not tailored to a specific backbone structure, but can be applied to any backbone with feature maps at different scales.
For this purpose, an additional grid rendering with individual weights for each scale present in the backbone generates feature maps from the input point cloud.
This is illustrated with orange framed grids in \fig{fig:network_architecture} for the backbone considered in this work.
In addition, radar point clouds tend to be sparse and the elevation information can be unprecise.
Thus, we apply a 2D BEV projection to the point cloud instead of 3D voxelization.

In the present case, the initial feature map with scale $s_0$ is successively downsampled by a factor of two during each stage of the residual network by applying strided convolutions.
Consequently, feature maps with spatial scales $s_i \in \{s_0,2 s_0, 4 s_0, 8 s_0\}$ are processed within the backbone.
For each of these scales, a grid with corresponding resolution is rendered directly from the point cloud in order to learn scale-aware features.
Subsequently, the resulting multi-scale feature maps are incorporated into the detection backbone by concatenation at the respective stage.

Generally, given an input which covers a certain area with a given resolution, the backbone downsamples the input to different scales.
For each spatial scale in the backbone, the feature map dimensions are typically divided (and are divisible) by an integer factor, which is equivalent to multiplying the resolution by this same factor.
In other words, for each spatial scale, the spatial dimensions should be divisible by the resolution to ensure perfect alignment.
In our case, the spatial dimensions should be divisible by $16s_0$ (including the last scale of the backbone).

Any grid rendering method that transforms irregular point cloud data into a regular grid representation can be used in this context, but we focus on the PointPillars and KPBEV grid rendering in this work.

\subsection{Multi-scale KPPillarsBEV architecture}

\begin{table*}[tb]
	\centering
	 \begin{minipage}{0.66\linewidth}
	 	\flushleft
	 	\vspace{0.1cm}
	 	\begin{tikzpicture}

\pgfdeclarelayer{background}
\pgfsetlayers{background,main}

\def\ypad{1.}
\def\xpad{0.4}

\definecolor{point_color}{RGB}{252,98,3}

\tikzstyle{arrow}=[thick, -stealth, rounded corners=3pt]
\tikzstyle{datatype}=[midway, align=center, font=\footnotesize, fill=white, inner sep=0cm]
\tikzstyle{stage}=[draw, thick, rounded corners, fill=white]
\tikzstyle{stagelabel} = [font=\small, fill=white, inner sep=0cm, align=center]
\tikzstyle{optionalstage}=[stage, dashed]
\tikzstyle{defaultsize}=[minimum width=2.7cm, minimum height=2.2cm]
\tikzstyle{layer}=[draw, rounded corners, font=\small, inner sep=0.1cm]

\newcommand{\pointcloud}[1] {
	\node at (#1) {
		\begin{tikzpicture}[scale=.4]
		\fill[point_color] (.6,.7) circle (.35em);
		\fill[point_color] (0,0) circle (.35em);
		\fill[point_color] (-.1,.6) circle (.35em);
		\fill[point_color] (-.5,-.4) circle (.35em);
		\fill[point_color] (0,-.4) circle (.35em);
		\fill[point_color] (.7,-.2) circle (.35em);
		\end{tikzpicture}	
	};
}

\newcommand{\grid}[1] {
	\node at (#1) {
		\begin{tikzpicture}[scale=.15]
		\begin{scope}[yslant=0.5,xslant=-1.]
		\draw[fill=white] (0.,0.) rectangle (3.,3.);
		\fill[point_color, opacity=0.2] (0,0) rectangle (1,1);
		\fill[point_color, opacity=0.4] (2,0) rectangle (3,1);
		\fill[point_color] (1,1) rectangle (2,2);
		\fill[point_color, opacity=0.5] (0,2) rectangle (1,3);
		\fill[point_color, opacity=0.7] (1,2) rectangle (2,3);
		
		\draw[step=1.,black,fill=white] (0., 0.) grid (3,3);
		\end{scope}
		\end{tikzpicture}	
	};
}

\node (preprocessing) [optionalstage, minimum width=2.2cm, minimum height=1cm] {};
\node [stagelabel] at (preprocessing.north) {preprocessing};
\node (kpconv) [layer] at ($(preprocessing)-(6pt, 0)$) {KPConv};
\node (n) [circle,draw=black, fill=black, text=white, inner sep=0pt,minimum size=5pt] at ($(kpconv.east)+(0.15,0)$) {$\mathbf{3\times}$};

\node (split) [shape=circle, fill=black, minimum size=0.1cm, inner sep=0cm] at ($(preprocessing.west)+(-\xpad,-\ypad)$) {};
\node (merge) [shape=diamond, draw] at ($(preprocessing.east)+(\xpad,-\ypad)$) {};

\node (gridrendering) [stage, defaultsize] at ($(merge)+(2.8,0)$) {};
\node [stagelabel] at (gridrendering.north) {grid rendering};
\node [stagelabel] at ($(gridrendering.north)+(0,0.4)$) {(multi-scale)};
\draw ($(gridrendering.west)+(0.2,0)$) -- node[midway,fill=white] {\small or} ($(gridrendering.east)-(0.2,0)$);
\begin{pgfonlayer}{background}
	\node (gr2) [optionalstage, defaultsize] at ($(gridrendering)+(0.3,0.5)$) {};
	\node (gr1) [optionalstage, defaultsize] at ($(gridrendering)+(0.15,0.25)$) {};
\end{pgfonlayer}

\node (pointpillars) [layer, minimum width=2.4cm] at ($(gridrendering)+(0,0.5)$) {(a) PointPillars};
\node (KPBEV) [layer, minimum width=2.4cm] at ($(gridrendering)-(0,0.5)$) {(b) KPBEV};

\node (backbone) [stage, align=center, inner sep=0.2cm] at ($(gridrendering)+(3.9,0)$) {Backbone\\and heads};

\draw[arrow] (split) -- (merge);
\draw[arrow, dashed] (split) |- (preprocessing.west);
\draw[arrow, dashed] (preprocessing.east) -| (merge);

\draw[arrow] (merge) -- node [datatype, xshift=-0.05cm] (labelmg) {point\\cloud} (gridrendering.west);
\pointcloud{$(labelmg)+(0,0.7)$}

\draw[arrow] ($(split)-(1.3,0)$) -- node [datatype, xshift=-0.05cm] (labelin) {point\\cloud} (split);
\pointcloud{$(labelin)+(0,0.7)$}

\draw[arrow] (gridrendering.east) -- node [datatype,xshift=0.12cm] (labelbb) {grid(s)} (backbone.west);
\grid{$(labelbb)+(0,0.5)$}

\end{tikzpicture}
	 	\vspace{0.1cm}
	 \end{minipage}\hfill\vline\hfill
	 \begin{minipage}{0.29\linewidth}
	 	\flushright
	 	\begin{tabular}{@{}lcc@{}}
	\toprule
	\textbf{Method} & \textbf{\begin{tabular}[c]{@{}c@{}}Pre-\\ proc.\end{tabular}} & \textbf{\begin{tabular}[c]{@{}c@{}}Grid\\ rendering\end{tabular}} \\ \midrule
	\vspace{0.1cm}PointPillars \cite{Lang2019}& \ding{55} & (a) \\
	\vspace{0.1cm}KPPillars \cite{Ulrich2022} & \ding{51} & (a) \\
	\vspace{0.1cm}KPBEV (ours) & \ding{55} & (b) \\
	KPPillarsBEV (ours) & \ding{51} & (b) \\ \bottomrule
\end{tabular}
	 \end{minipage}
	
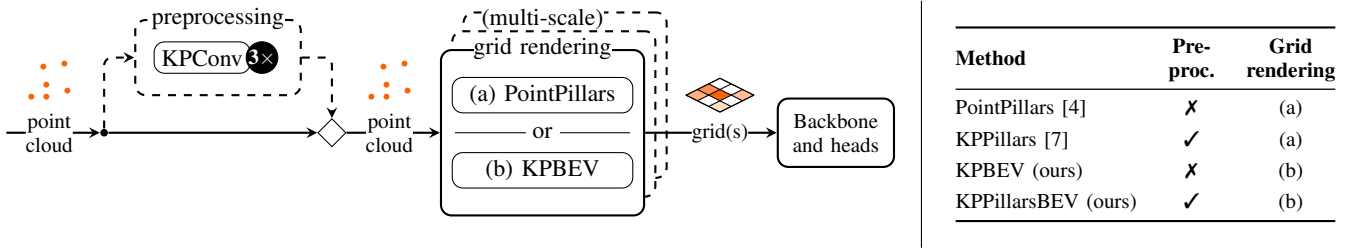
\captionof{figure}{Overview of the different architectures considered in this work in order to investigate the benefits of multi-scale and KPBEV grid rendering. All architectures share the same structure for backbone and detection heads and are evaluated with single-scale and our new multi-scale grid rendering respectively. Optionally, the point cloud is first preprocessed with a point-based network consisting of three KPConv layers, as in \cite{Ulrich2022}. Subsequently, the point cloud is rendered into one or multiple grids using either a PointPillars or our novel KPBEV feature encoder, before being fed into the backbone.}
	\label{fig:architectures}
\end{table*}

We propose the multi-scale KPPillarsBEV radar object detection architecture which brings together KPBEV grid rendering, multi-scale grid rendering and point pre-processing.
We also examine different variations of multi-scale KPPillarsBEV with and without the proposed methods as illustrated in \fig{fig:architectures}.
This enables us to perform an ablation study and quantify the benefits of each contribution.
Moreover, these variations offer different tradeoffs between detection performance and complexity.

The multi-scale KPPillarsBEV architecture first preprocesses the input point cloud with three sequential KPConv layers as in \cite{Ulrich2022} and enriches the point cloud with additional scale-agnostic features.
Then, following the proposed multi-scale grid rendering formulation, the resulting point cloud is rendered to a corresponding grid with a KPBEV module for each scale of the backbone.
As a result, the point is rendered to the grid with scale-specific features as each KPBEV module has its own set of weights.
Finally, the resulting feature maps are processed by a ResNet backbone, a feature pyramid network and detection heads as illustrated in \fig{fig:network_architecture}.
The predicted oriented bounding boxes are then filtered with non-maximum suppression.

We evaluate different tradeoffs between detection performance and computational complexity with variants of multi-scale KPPillarsBEV as illustrated in \fig{fig:architectures}.
Specifically, we consider three dimensions: the presence of the KPConv preprocessing, the grid rendering method (PointPillars or KPBEV) and whether the grid rendering is multi- or single-scale.
Within this framing, the evaluated PointPillars \cite{Lang2019} architecture has no KPConv preprocessing and single-scale PointPillars grid rendering.
Similarly, KPPillars \cite{Ulrich2022} uses KPConv preprocessing but only has single-scale PointPillars grid rendering.
On the other hand, multi-scale KPPillarsBEV has preprocessing and multi-scale KPBEV grid rendering.
As a result, it effectively combines scale-agnostic and scale-specific processing and minimizes information loss from the point cloud to the grid.
Indeed, we find in our experiments that the combination of preprocessing, KPBEV and multi-scale grid rendering significantly improves detection performance compared to other variants.

All architectures share the same backbone structure depicted in \fig{fig:network_architecture}, that consists of a residual network (ResNet) \cite{He2016} and feature pyramid network (FPN) \cite{Lin2017} and is similar to \cite{Yang2019}.
The generation of oriented bounding boxes (OBBs) is performed by convolutional detection heads that generate object proposals for specific class groups, such as vulnerable road users (VRUs) or car-like objects.

In addition, when KPBEV grid rendering is performed in a multi-scale manner, it is essential that the influence radius of the KPBEV is coherent with the grid resolution.
If the influence radius is kept constant at all spatial scales, the neighborhood may become smaller than the cell size for coarse grids.
Consequently, we propose to use an adaptive influence radius $\rho_{k,i}=(s_i/s_0)\rho_{k,0}$ for each corresponding scale $s_i$.
This is crucial in order for multi-scale KPBEV to significantly outperform single-scale KPBEV.
\section{Experiments}
\label{sec:experiments}

\subsection{Experimental setup}

We evaluate the proposed methods on the nuScenes validation set \cite{Caesar2020}, which contains multi-modal sensor data of urban driving scenarios captured in Boston and Singapore.
To improve the comparability of our work, we use the official evaluation toolkit and metrics \cite{Caesar2020}. 
In the nuScenes benchmark, average precision (AP) metrics are provided for different matching thresholds between ground truth and predictions (0.5, 1, 2 and 4 meters) to quantify the detection performance.
Additionally, the mean average precision (mAP) over the different matching thresholds is computed.
For a detailed explanation, we refer the reader to \cite{Caesar2020}.

In our evaluation we focus on the AP4.0, \ie the AP with a matching threshold of 4\,meters, and the mAP for class \textit{car}.
Although the models are also trained to detect other classes such as \textit{pedestrian}, the performance for these classes is quite low. 
This was already observed in previous radar literature \cite{Nobis2021,Svenningsson2021,Ulrich2022} and is mostly due to the limited resolution of radars in nuScenes, which results in many objects having no or only very few reflections.
Furthermore, the sparsity and the lack of elevation in nuScenes' radar data impedes classification for some objects, leading to confusion between classes such as \textit{bus} and \textit{truck}.

Besides detection performance, we assess computational complexity of the architectures in terms of GPU runtime and peak memory utilization on a \texttt{NVIDIA V100 SMX2} GPU with the TensorFlow \cite{Abadi2016} profiler.

\subsection{Models}

\begin{table*}[ht!]
	\newcommand\modelspacing{0.154cm}
	\centering
	\caption{Benchmark of the different architectures for class \textit{car} on the nuScenes validation set.}
	\figuredesc{The proposed multiscale KPPillarsBEV significantly outperforms \cite{Lang2019,Ulrich2022} in terms of AP4.0 and mAP at a lower but still acceptable frame rate. Single-scale KPBEV provides a significant performance boost compared to single-scale PointPillars with no additional computational cost. As expected, using multi-scale grids improves performance of all methods.\vspace{0.25cm}}
	\label{tab:benchmark_nuscenes}
	\begin{tabular}{@{}lccccc@{}}
	\toprule
	\textbf{Method} & \textbf{Multi-scale} & \textbf{AP4.0 [\%] $\uparrow$} & \textbf{mAP [\%] $\uparrow$} & \textbf{FPS [Hz] $\uparrow$} & \textbf{GPU Mem. [MB] $\downarrow$} \\ \midrule
	\multirow{2}{*}{PointPillars   \cite{Lang2019} (baseline)} & \ding{55} & 38.31 & 22.08 & 86.96 & \textbf{89.73} \\
	& \ding{51} & 38.70 & 22.44 & 76.28 & 98.32\vspace{\modelspacing} \\
	\multirow{2}{*}{KPPillars   \cite{Ulrich2022}} & \ding{55} & 40.80 & 24.01 & 78.13 & 99.79 \\
	& \ding{51} & 41.47 & 24.37 & 68.78 & 100.23\vspace{\modelspacing} \\
	\multirow{2}{*}{KPBEV (ours)} & \ding{55} & 41.21 & 24.50 & \textbf{87.11} & 91.82 \\
	& \ding{51} & 42.27 & 25.26 & 67.20 & 108.77\vspace{\modelspacing} \\
	\multirow{2}{*}{KPPillarsBEV (ours)} & \ding{55} & 41.84 & 24.95 & 74.85 & 103.18 \\
	& \ding{51} & \textbf{43.68} & \textbf{26.42} & 60.02 & 111.18 \\\hline
\end{tabular}
\end{table*}

\begin{figure}[ht!]
	\centering
	 \includegraphics[width=.8\columnwidth]{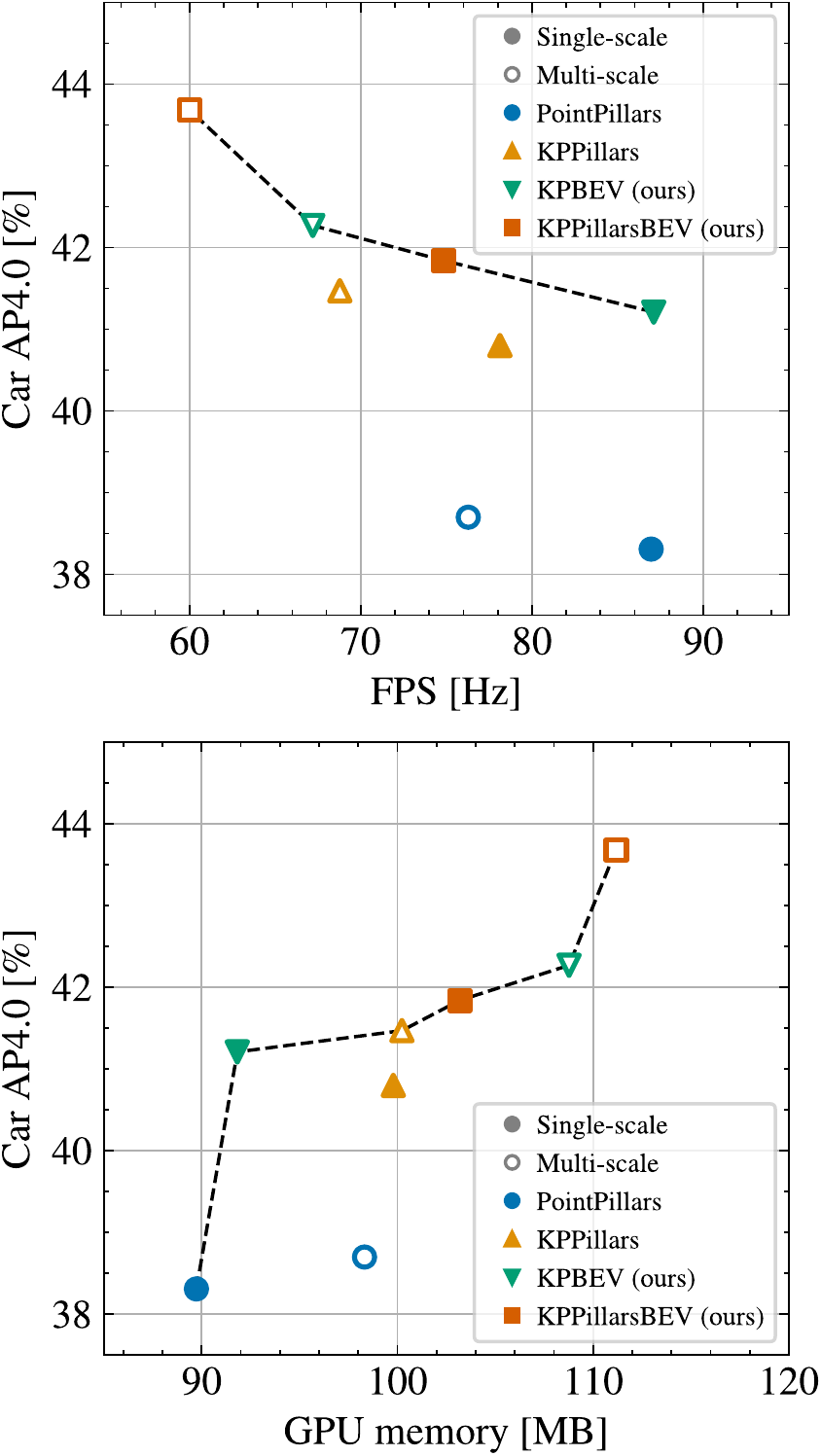}
	 \caption{Number of frames per second (FPS) and peak memory usage in relation to the detection performance (AP4.0) for class \textit{car} on nuScenes achieved by the different methods investigated in this work. The black dashed lines indicate the respective Pareto front.}
	\label{fig:pareto_fronts}
\end{figure}

We investigate the influence of multi-scale grid rendering and the novel KPBEV feature encoder.
To this end, we consider four different evaluation settings (see \fig{fig:architectures}):
\begin{itemize}
	\item PointPillars: the baseline setting which uses a point cloud encoder based on a simplified PointNet as in \cite{Lang2019} to convert the point cloud into a BEV feature map.
	\item KPPillars: state-of-the-art radar setting \cite{Ulrich2022}, that uses point-based preprocessing consisting of three KPConv layers before applying the PointPillars encoder.
	\item KPBEV (ours): low complexity setting that uses the KPBEV feature encoder (see \fig{fig:kpbev}) proposed in this work instead of PointPillars.
	\item KPPillarsBEV (ours): high performance setting that combines the point-based preprocessing of KPPillars with KPBEV grid rendering.
\end{itemize}
Each of these settings is evaluated with single- and our proposed multi-scale grid rendering.


\subsection{Experiment parameters}

The network input is a radar point cloud that is temporally aggregated over five consecutive measurements. 
Each point carries the features $\{x, v_r, \sigma, \Delta t\}$, where $x\in\mathbb{R}^2$ are the 2D Cartesian coordinates, $v_r\in\mathbb{R}$ is the ego-motion compensated radial velocity, $\sigma\in\mathbb{R}$ is the radar cross section and $\Delta t\in\mathbb{R}$ is the timestamp difference \wrt the most recent timestamp.
Object detection is then performed on a grid ranging from -60 to 60\,meters in both the x- and y-direction with an initial cell size of $s_0=0.5\,m$.
All layers preceding the detection backbone use $F_\textnormal{out}=64$ channels to process the input point cloud.
For KPBEV an influence radius of $\rho_{k,0}=0.6\,m$ is used for the initial scale $s_0$, while for KPPillars and KPPillarsBEV $\rho_{k,0}=1\,m$ is used as proposed by \cite{Ulrich2022}.
For KPBEV and KPPillarsBEV, we use the proposed adaptive influence radius scheme in the multi-scale variants.

We run 10 trials per model variant and calculate the average of metrics to account for stochasticity during training. 
Each model is trained for 50 epochs with a batch size of 24. 

\subsection{Results}

\tab{tab:benchmark_nuscenes} shows the quantitative results of the different methods on the nuScenes validation set.
For all architectures, we observe that multi-scale grid rendering always improves the detection performance over single-scale grid rendering.
The absolute improvements in AP4.0 and mAP range from $0.39\,\%$ and $0.36\,\%$ for the baseline to up to $1.82\,\%$ and $1.47\,\%$ for KPPillarsBEV.
Furthermore, the benefits of multi-scale grid rendering are greater for models with KPBEV grid rendering instead of PointPillars.
Hence, we conclude that KPBEV is able to learn more expressive and scale-aware features, due to the improved cell-wise feature aggregation from neighboring points.


This is also reflected when comparing single-scale PointPillars and KPBEV.
KPBEV achieves a significant performance increase of $2.90\,\%$ in AP4.0 and $2.42\,\%$ in mAP over the baseline, without increasing the inference time. 
Additionally, KPBEV outperforms KPPillars, the previous state-of-the-art radar model, in terms of both detection performance and computational efficiency.

\begin{figure*}[ht!]
	\centering
	\input{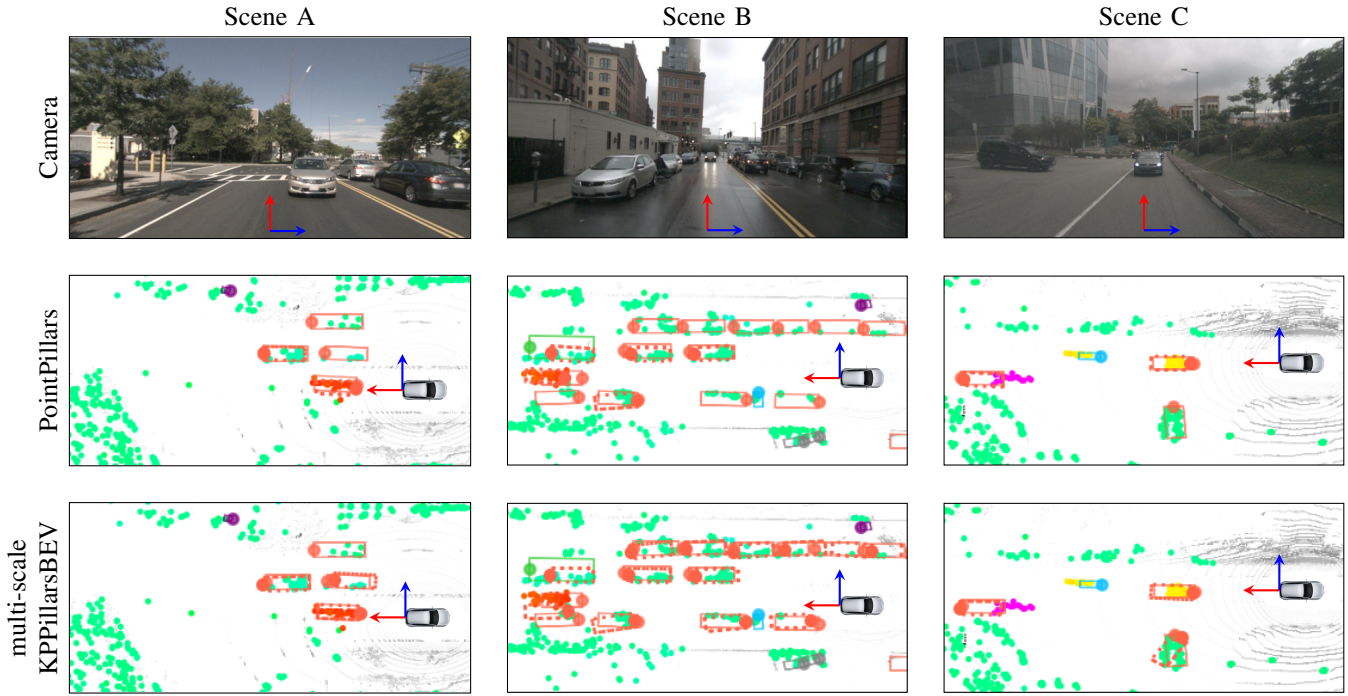}
	\caption{Images of the rear camera and corresponding qualitative results of PointPillars and multi-scale KPPillarsBEV in bird's-eye perspective for different scenes on the nuScenes validation set. The results show the radar point cloud (colored points, where the color encodes radial velocity), lidar point cloud (grey points), ground truth boxes (solid rectangles) and predicted boxes (dotted rectangles). The qualitative results illustrate the improved detection performance of KPPillarsBEV, as it precisely detects cars that are missed by the baseline (PointPillars), \eg the parked cars in Scene B.}
	\label{fig:qualitative_results}
\end{figure*}

The main reason for the efficiency of KPBEV is, that KPConvs are only applied to anchor points, \ie once per non-empty grid cell.
In contrast, during the point-based preprocessing of KPPillars the convolutions are applied to each input point.
For instance, the number of KPConvs is reduced by $38\,\%$ for a grid resolution of $0.5\,m$ as there is that many less anchor points than input points.

This property also alleviates the use of larger convolution radii in the multi-scale configuration.
As previously described, the convolution radius is increased adaptively for coarser grids.
Consequently, the number of neighboring points considered during the feature aggregation at each anchor point grows.
At the same time, the number of non-empty cells and thus the number of necessary convolutions decreases as cells become larger.
However, using KPBEV at multiple scales still decreases the frame rate by about 20\,\%.

The trade-off between detection performance and computational complexity is visualized by the Pareto fronts in \fig{fig:pareto_fronts}.
In summary, both multi-scale and KPBEV grid rendering benefit the performance of radar object detection networks.
The modularity of these approaches help to design suitable, Pareto-optimal architectures.
If maximum detection performance is desired, the proposed multi-scale KPPillarsBEV architecture is the best option and achieves an AP4.0 of $43.68\,\%$ and mAP of $26.42\,\%$ in our experiments.
The strong detection performance of this method is also illustrated in \fig{fig:qualitative_results}, which shows qualitative results for several scenes in the nuScenes validation set in comparison to the PointPillars baseline.
In a scenario with limited compute resources, the single-scale KPBEV architecture may be preferable, as it significantly outperforms the baseline while maintaining the same inference speed.




\section{Conclusion}

Grid-based approaches \cite{Ulrich2022} are currently the state-of-the-art for radar object detection.
However, current approaches tend to lose information during the grid rendering step due to aggregation and discretization.
To address this, we propose a novel multi-scale KPPillarsBEV architecture which brings together KPConv preprocessing, a novel KPBEV grid rendering and a general multi-scale grid rendering formulation.

Specifically, we introduced two new methods to improve grid rendering for radar-based object detection networks in a principled manner: KPBEV grid rendering and general multi-scale grid rendering.
The proposed KPBEV grid rendering based on kernel point convolutions has greater descriptive power than previous encoders which rely on simplified PointNets and mitigates information loss due to aggregation and discretization.
Our experiments show that single-scale KPBEV significantly improves detection performance compared to the PointPillars baseline at no additional complexity cost.

Furthermore, we propose a general multi-scale grid rendering formulation which performs grid rendering for each spatial scale with any grid rendering method and incorporates scale-aware features into the detection backbone.
We show experimentally that multi-scale grid rendering improves the detection performance for all considered architectures.
Moreover, the performance gains are the largest when used in combination with the proposed KPBEV grid rendering.

The proposed multi-scale KPPillarsBEV architecture outperforms the current state of the art for radar object detection on nuScenes.
It achieves an AP4.0 of $43.68\,\%$ and mAP of $26.42\,\%$ for class \textit{car}, while running at 60\,FPS on a \texttt{NVIDIA V100 SMX2} GPU, and outperforms the baseline PointPillars by 5.37\,\% AP4.0.



\section{Acknowledgements}

This work was supported by the German Federal Ministry
of Education and Research, project ZuSE-KI-AVF under grant
\no 16ME0062.

\bibliographystyle{IEEEtran}
\bibliography{references}

\end{document}